\documentclass{bmvc2k}
\usepackage{amsmath,amssymb}

\title{Speeding up Convolutional Neural Networks with Low Rank Expansions}

\addauthor{Max Jaderberg}{max@robots.ox.ac.uk}{1}
\addauthor{Andrea Vedaldi}{vedaldi@robots.ox.ac.uk}{1}
\addauthor{Andrew Zisserman}{az@robots.ox.ac.uk}{1}

\addinstitution{
 Visual Geometry Group\\
 University of Oxford\\
 Oxford, UK
}

\runninghead{Jaderberg, Vedaldi, and Zisserman}{Speeding up Convolutional ...}

\def\eg{\emph{e.g}\bmvaOneDot}
\def\ie{\emph{i.e}\bmvaOneDot}

\def\etal{\emph{et al}\bmvaOneDot}

\renewcommand{\paragraph}[1]{\par\medskip\noindent{\bf #1}}

\begin{document}
\maketitle
\begin{abstract}
The focus of this paper is speeding up the evaluation of convolutional
neural networks. While delivering impressive results across a range of
computer vision and machine learning tasks, these networks are
computationally demanding, limiting their deployability.
Convolutional layers generally consume the bulk of the
processing time, and so in this work we present two simple schemes for
drastically speeding up these layers. This is achieved by exploiting
cross-channel or filter redundancy to construct a low rank basis of filters
that are rank-1 in the spatial domain. Our methods are architecture
agnostic, and can be easily applied to existing CPU and GPU
convolutional frameworks for tuneable speedup performance. We
demonstrate this with a real world network designed for scene text
character recognition, showing a possible 2.5$\times$ speedup with no loss in accuracy, and 4.5$\times$ speedup with less than
1\% drop in accuracy, still achieving state-of-the-art on standard benchmarks.
\end{abstract}

\section{Introduction}\label{sec:intro}

Many applications of machine learning, and most recently computer vision, have been disrupted by the use of convolutional neural networks (CNNs). The combination of an end-to-end learning system with minimal need for human design decisions, and the ability to efficiently train large and complex models, have allowed them to achieve state-of-the-art performance in a number of benchmarks~\cite{Krizhevsky12,Taigman14,Toshev13,Alsharif13,Goodfellow13,Sermanet13,Razavian14}. However, these high performing CNNs come with a large computational cost due to the use of chains of several convolutional layers, often requiring implementations on GPUs~\cite{Krizhevsky12,Jia13caffe} or highly optimized distributed CPU architectures~\cite{Vanhoucke11} to process large datasets. The increasing use of these networks for detection in sliding window approaches~\cite{Sermanet13,Farabet12, Oquab13} and the desire to apply CNNs in real-world systems means the speed of inference becomes an important factor for applications. In this paper we introduce an easy-to-implement method for significantly speeding up pre-trained CNNs requiring minimal modifications to existing frameworks. There can be a small associated loss in performance, but this is tunable to a desired accuracy level. For example, we show that a 4.5$\times$ speedup can still give state-of-the-art performance in our example application of character recognition.

While a few other CNN acceleration methods exist, our {\bf key insight is to exploit the redundancy that exists between different feature channels and filters}~\cite{Denil13}. We contribute two approximation schemes to do so (Sect.~\ref{sec:separable}) and two optimization methods for each scheme (Sect.~\ref{sec:opt}). Both schemes are orthogonal to other architecture-specific optimizations and can be easily applied to existing CPU and GPU software. Performance is evaluated empirically in Sect.~\ref{sec:exp} and results are summarized in Sect~\ref{sec:conc}.

\paragraph{Related work.} 
There are only a few general speedup methods for CNNs. Denton~\etal~\cite{Denton2014} use low rank approximations and clustering of filters achieving 1.6$\times$ speedup of single convolutional layers (not of the whole network) with a 1\% drop in classification accuracy. Mamalet~\etal~\cite{Mamalet12} design the network to use rank-1 filters from the outset and combine them with an average pooling layer; however, the technique cannot be applied to general network designs. Vanhoucke~\etal~\cite{Vanhoucke11} show that 8-bit quantization of the layer weights can result in a speedup with minimal loss of accuracy. Not specific to CNNs, Rigamonti~\etal~\cite{Rigamonti13} show that multiple image filters can be approximated by a shared set of separable (rank-1) filters, allowing large speedups with minimal loss in accuracy.

Moving to hardware-specific optimizations, \texttt{cuda-convnet}~\cite{Krizhevsky12} and \texttt{Caffe}~\cite{Jia13caffe} show that highly optimized CPU and GPU code can give superior computational performance in CNNs. \cite{Mathieu13} performs convolutions in the Fourier domain through FFTs computed efficiently over batches of images on a GPU. Other methods from~\cite{Vanhoucke11} show that specific CPU architectures can be taken advantage of, \eg by using SSSE3 and SSSE4 fixed-point instructions and appropriate alignment of data in memory. Farabet~\etal~\cite{Farabet11} show that using bespoke FPGA implementations of CNNs can greatly increase processing speed.

To speed up test-time in a sliding window context for a CNN,~\cite{Iandola14} shows that multi-scale features can be computed efficiently by simply convolving the CNN across a flattened multi-scale pyramid. Furthermore search space reduction techniques such as selective search~\cite{Vandesande11} drastically cut down the number of times a full forward pass of the CNN must be computed by cheaply identifying a small number of candidate object locations in the image.

Note, the
methods we proposed are not specific to any processing architecture and can be
combined with many of the other speedup methods given above.

\nopagebreak
\section{Filter Approximations}\label{sec:separable}
\nopagebreak

Filter banks are used widely in computer vision as a method of feature
extraction, and when used in a convolutional manner, generate
\emph{feature maps} from input images. For an input $x \in
\mathbb{R}^{H \times W}$, the set of output feature maps $Y=\{y_1,
y_2, \dots, y_N\}$, $y_n \in \mathbb{R}^{H' \times W'}$ are generated
by convolving $x$ with $N$ filters $F=\{f_i\}~\forall i\in[1\dots N]$ such
that $y_i=f_i * x$. The collection of filters $F$ can be learnt, for
example, through dictionary learning
methods~\cite{Kavukcuoglu10,Lee09,Rigamonti11} or CNNs, and are
generally full rank and expensive to convolve with large images. Using
a direct implementation of convolution, the complexity of convolving a
single channel input image with a bank of $N$ 2D filters of size $d
\times d$ is $\mathcal{O}(d^2NH'W')$.
We next introduce our method for accelerating this computation that
takes advantage of the fact that {\bf there
exists significant redundancy \emph{between} different filters and
feature channels}.

One way to exploit this redundancy is to approximate the filter set by
a linear combination of a smaller basis set of $M$
filters~\cite{Rigamonti13,Song12,Song13}. The basis filter set
$S=\{s_i\}~\forall i\in[1\dots M]$ is used to generate basis feature
maps which are then linearly combined such that $y_i \simeq
\sum_{k=1}^{M} a_{ik}s_k*x $. This can lead to a speedup in feature map
computation as a smaller number of filters need be convolved with the
input image, and the final feature maps are composed of a cheap linear
combination of these. The complexity in this case is
$\mathcal{O}((d^2M + MN)H'W')$, so a speedup can be achieved if $M <
\frac{d^2N}{d^2 + N}$. 

As shown in Rigomonti~\etal~\cite{Rigamonti13}, further speedups can be achieved by choosing the filters in the approximating basis to be rank-1 and so making individual convolutions \emph{separable}. This means that each basis filter can be decomposed in to a sequence of horizontal and vertical filters $s_i * x = v_i * (h_i *x)$ where $s_i \in \mathbb{R}^{d \times d}$, $v_i \in \mathbb{R}^{d \times 1}$, and $h_i \in \mathbb{R}^{1 \times d}$. Using this decomposition, the convolution of a separable filter $s_i$ can be performed in $\mathcal{O}(2dH'W')$ operations instead of $\mathcal{O}(d^2H'W')$. 


The separable filters of~\cite{Rigamonti13} are a low-rank
approximation, but performed in the {\em spatial} filter dimensions. Our key
insight is that in CNNs substantial speedups can be achieved by
also exploiting the cross-channel redundancy to perform low-rank
decomposition in the {\em channel} dimension as well. 
We explore both of these low-rank approximations in the sequel.

Note that the FFT~\cite{Mathieu13} could be used as an alternative
speedup method to accelerate individual convolutions in combination
with our low-rank cross-channel decomposition scheme. However,
separable convolutions have several practical advantages: they are
significantly easier to implement than a well tuned FFT
implementation, particularly on GPUs; they do not require feature maps
to be padded to a special size, such as a power of two as
in~\cite{Mathieu13}; they are far more memory efficient; and, they yield a good speedup for small image
and filter sizes too (which can be common in CNNs), whilst FFT
acceleration tends to be better for large filters due to the overheads
incurred in computing the FFTs.

\subsection{Approximating Convolutional Neural Network Filter Banks}\label{sec:cnns}
CNNs are obtained by stacking multiple layers of convolutional filter banks on top of each other, followed by a non-linear response function. Each filter bank or \emph{convolutional layer} takes an input which is a feature map $z_i(u,v)$ where $(u,v)\in\Omega_i$ are spatial coordinates and  $z_i(u,v) \in \mathbb{R}^C$ contains $C$ scalar features or \emph{channels} $z_i^c(u,v)$. The output is a new feature map $z_{i+1} \in \mathbb{R}^{H' \times W' \times N}$ such that
$
z^n_{i+1} = h_i(W_{in} \ast z_i + b_{in})~\forall n \in [1\dots N],
$ where $W_{in}$ and $b_{in}$ denote the $n$-th filter kernel and bias
respectively, and $h_i$ is a non-linear activation function such as
the \emph{Rectified Linear Unit} (ReLU) $h_i(z) =
\max\{0,z\}$. Convolutional layers can be intertwined with
\emph{normalization}, \emph{subsampling}, and \emph{pooling
layers} which build translation invariance in local neighbourhoods. Other layer types are possible as well, but generally the convolutional ones are the most expensive. The
process starts with $z_1=x$, where $x$ is the input image, and ends by, for example, connecting the last feature map to a logistic regressor in the case of classification. All the parameters of
the model are jointly optimized to minimize a loss
over the training set using Stochastic Gradient Descent (SGD) with back-propagation.

\begin{figure}[t]
\raggedright
\begin{minipage}[b]{0.5\textwidth}\centering
\includegraphics[scale=0.24]{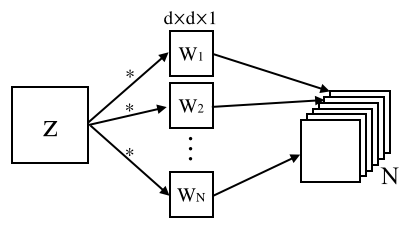}(a)
\end{minipage}
\includegraphics[width=16em]{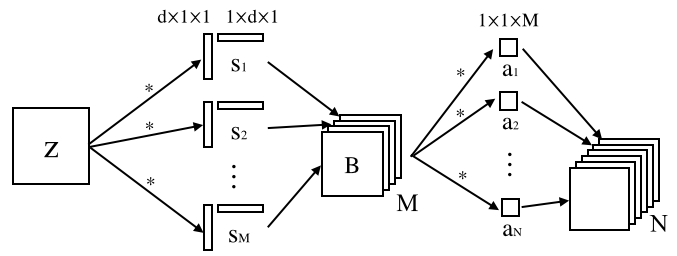}(b)
\begin{minipage}[b]{0.5\textwidth}\caption{\small 
(a) The original convolutional layer acting on a single-channel input
\ie C=1. (b) The approximation to that layer using the method of
Scheme 1. (c) The approximation to that layer using the method of
Scheme 2. Individual filter dimensions are given above the filter
layers.}
\label{fig:scheme1}
\label{fig:scheme2}
\end{minipage}
\includegraphics[width=16em]{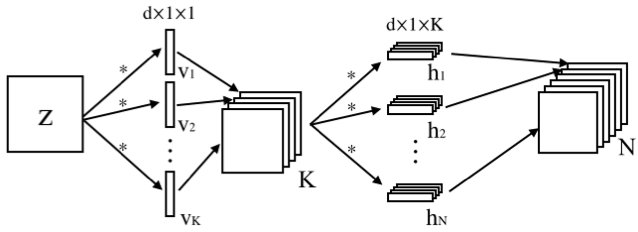}(c)
\end{figure}

The $N$ filters $W_{n}$ learnt for each layer (for convenience we drop the layer subscript $i$) are full rank, 3D filters with the same depth as the number of channels of the input, such that $W_{n}(u,v) \in \mathbb{R}^C$. For example, for a 3-channel color image input, $C=3$. The convolution $W_{n} * z$ of a 3D filter $W_{n}$ with the 3D image $z$ is the 2D image $W_{n} * z = \sum_{c=1}^C W_{n}^c * z^c$, where $W_{n}^c \in \mathbb{R}^{d \times d}$ is a single channel of the filter. This is a sum of 2D convolutions so we can think of each 3D filter as being a collection of 2D filters, whose output is collapsed to a 2D signal. However, since $N$ such 3D filters are applied to $z$, the overall output is a new 3D image with $N$ channels. This process is illustrated for the case $C=1, N>1$ in Fig.~\ref{fig:scheme1}~(a). The resulting computational cost for a convolutional layer with $N$ filters of size $d \times d$ acting on $C$ input channels is $\mathcal{O}(CNd^2H'W')$.

We now propose two schemes to approximate a convolutional layer of a CNN to reduce the computational complexity and discuss their training in Sec.~\ref{sec:opt}. Both schemes follow the same intuition: that CNN filter banks can be approximated using a low rank
 basis of filters that are separable in the spatial domain.

\paragraph{Scheme 1.} 
The first method for speeding up convolutional layers is to directly apply the method suggested in Sect.~\ref{sec:separable} to the filters of a CNN~(Fig.~\ref{fig:scheme1} (b)). As described above, a single convolutional layer with $N$ filters $W_{n} \in \mathbb{R}^{d \times d \times C}$ requires evaluating $NC$ 2D filters $F = \{W_{n}^c \in\mathbb{R}^{d\times d}: n \in [1\dots N], c \in [1\dots C]\}$. Note that there are $N$ filters $\{W_n^c:n\in [1\dots N]\}$ operating on each input channel $z^c$. These can be approximated as linear combinations of a basis of $M < N$ (separable) filters $S^c=\{s_m^c : m \in [1\dots M]\}$ as $W_{n}^c \simeq \sum_{m=1}^M a_{n}^{cm}s^c_m$. Since convolution is a linear operator, filter reconstruction and image convolution can be swapped, yielding the approximation $W_n * z = \sum_{c=1}^{C} W_n^c * z^c \simeq \sum_{c=1}^C \sum_{m=1}^M a_{n}^{cm} (s_m^c * z^c)$. To summarize, the direct calculation involves computing $NC$ 2D filters $W_n^c * z^c$ with cost $O(NC d^2 H'W')$, while the approximation involves computing $MC$ 2D  filters $s_m^c*z^c$ with cost $O(MC (d^2 + N) H'W')$ -- the additional $MC NH'W'$ term accounting for the need to recombine the basis response linearly. Due to the second term, the approximation is efficient only if $M \ll d^2$, \ie if the number of filters in the basis is less than the filter area.

The first cost term $CM d^2 H'W'$ would also suggest that efficiency requires the condition $M \ll N$; however, this can be considerably ameliorated by using \emph{separable filters} in the basis. In this case the approximation cost is reduced to $O(MC (d + N )H'W')$; together with the former condition, Scheme 1 is then efficient if $M \ll d\min\{d,N\}$.

Note that this scheme uses $C$ filter basis $S^1,S^2,\dots,S^C$ as each operates on a different input channel. In practice, we choose $S^1=S^2=\dots=S^C=S$ because empirically there is no actual gain in performance and a single channel basis is simpler and more compact.


\paragraph{Scheme 2.}
Scheme 1 focuses on approximating 2D filters. As a consequence, each input channel $z^c$ is approximated by a particular basis of 2D separable filters. Redundancy among feature channels is exploited, but only in the sense of the $N$ \emph{output channels}. In contrast, Scheme 2 is designed to take advantage of both input and output redundancies by considering 3D filters throughout. The idea is simple: each convolutional layer is factored as a sequence of two regular convolutional layers but with rectangular (in the spatial domain) filters, as shown in Fig.~\ref{fig:scheme2} (c). The first convolutional layer has $K$ filters of spatial size $d \times 1$ resulting in a filter bank $\{v_k \in \mathbb{R}^{d \times 1 \times C}:  k \in [1\dots K]\}$ and producing output feature maps $V$ such that $V(u,v) \in \mathbb{R}^{K}$. The second convolutional layer has $N$ filters of spatial size $1 \times d$ resulting in a filter bank $\{h_n \in \mathbb{R}^{1 \times d \times K}:  n \in [1\dots N]\}$. Differently from Scheme 1, the filters operate on multiple channels simultaneously. The rectangular shape of the filters is selected to match a separable filter approximation. To see this, note that convolution by  one of the original filters $W_{n} * z = \sum_{c=1}^C W_n^c * z^c$ is approximated by 
\vspace{-0.5em}
\begin{equation}\label{eqn:scheme2}
W_{n} * z
\simeq h_n * V
= \sum_{k=1}^K h_n^k * V^k 
= \sum_{k=1}^K h_n^k * (v_k * z)
= \sum_{k=1}^K h_n^k * \sum_{c=1}^C v_k^c * z^c 
= \sum_{c=1}^C \left[\sum_{k=1}^K h_n^k * v_k^c \right] * z^c
\end{equation}
which is the sum of separable filters $h_n^k*v_k^c$. The computational cost of this scheme is $O(KCd H'W)$ for the first vertical filters and $O(NK d H'W')$ for the second horizontal filter. Assuming that the image width $W\gg d$ is significantly larger than the filter size, the output image width $W\approx W'$ is about the same as the input image width $W'$. Hence the total cost can be simplified to $O(K(N+C) d H'W')$. Compared to the direct convolution cost of $O(NC d^2 H'W')$, this scheme is therefore convenient provided that $K(N+C) \ll NC d$. For example, if $K$, $N$, and $C$ are of the same order, the speedup is about $d$ times.


In both schemes, we are assuming that the full rank original convolutional filter bank can be decomposed in to a linear combination of a set of separable basis filters. The difference between the schemes is how/where they model the interaction between input and output channels, which amounts to how the low rank channel space approximation is modelled. In Scheme 1 it is done with the linear combination layer, whereas with Scheme 2 the channel interaction is modelled with 3D vertical and horizontal filters inducing a summation over channels as part of the convolution.

\subsection{Optimization}\label{sec:opt}
This section deals with the details on how to attain the optimal separable basis representation of a convolutional layer for the schemes. The first method (Sec.~\ref{sec:filterrecon}) aims to reconstruct the original filters directly by minimizing filter reconstruction error. The second method (Sec.~\ref{sec:datarecon}) approximates the convolutional layer indirectly, by minimizing reconstruction error of the output of the layer.

\subsubsection{Filter Reconstruction Optimization}\label{sec:filterrecon}
The first way that we can attain the separable basis representation is to aim to minimize the reconstruction error of the original filters with our new representation. 

\paragraph{Scheme 1.}
The separable basis can be learnt simply by minimizing the $L_2$ reconstruction error of the original filters, whilst penalizing the nuclear norm $\|s_m\|_*$ of the filters $s_m$. In fact, the nuclear norm $\|s_m\|_*$ is a proxy for the rank of $s_m\in \mathbb{R}^{d\times d}$ and rank-1 filters are separable. This yields the formulation:
\begin{equation}
\underset{\{s_m\}, \{a_{n}\}} {\mathrm{min}}~\sum_{n=1}^N \sum_{c=1}^C \left\| W_{n}^c - \sum_{m=1}^M a_{n}^{cm}s_m \right\|_2^2 + \lambda \sum_{m=1}^M \| s_m \|_*.
\end{equation}
This minimization is biconvex, so given ${s_m}$ a unique $a_n$ can be found, therefore a minimum is found by alternating between optimizing $s_m$ and $a_n$. For full details of the implementation of this optimization see~\cite{Rigamonti13}.

\paragraph{Scheme 2.}
The set of horizontal and vertical filters can be learnt by explicitly minimizing the $L_2$ reconstruction error of the original filters. From~\eqref{eqn:scheme2} we can see that the original filter can be approximated by minimizing the objective function
\begin{equation}
\underset{\{h_n^k\}, \{v_k^c\}} {\mathrm{min}}~\sum_{n=1}^N \sum_{c=1}^C \left\| W_n^c - \sum_{k=1}^K h_n^k * v_k^c \right\|_2^2.
\end{equation}
This optimization is simpler than for Scheme 1 due to the lack of nuclear norm constraints, which we are able to avoid by modelling the separability explicitly with two variables. We perform conjugate gradient descent, alternating between optimizing the horizontal and vertical filter sets. 

\vspace{-2mm}
\subsubsection{Data Reconstruction Optimization}\label{sec:datarecon}

\begin{figure}[t]
\centering
\begin{tabular}{cccc}
\includegraphics[width=0.36\textwidth]{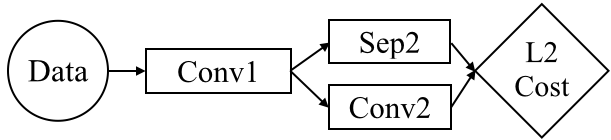}&
\includegraphics[width=0.44\textwidth]{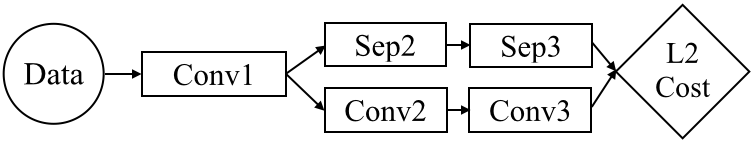}
\\
(a)&(b)
\end{tabular}
\caption{\small Example schematics of how to optimize separable basis approximation layers in a data reconstruction setting. (a) Approximating Conv2 with Sep2. (b) Approximating Conv3 with Sep3, incorporating the approximation of Conv2 as well.}
\label{fig:reconnet}
\end{figure}

The problem with optimizing the separable basis through minimizing original filter reconstruction error is that this does not necessarily give the most optimized basis set for the end CNN prediction performance. As an alternative, one can optimize a scheme's separable basis by aiming to reconstruct the \emph{outputs} of the original convolutional layer given training data. For example, for Scheme 2 this amounts to
\vspace{-4mm}
\begin{equation}
\underset{\{h_n^k\}, \{v_k^c\}} {\mathrm{min}}~\sum_{i=1}^{|X|} \sum_{n=1}^N \left\| W_{n}*\Phi_{l-1}(x_i) - \sum_{c=1}^C \sum_{k=1}^K h_n^k * v_k^c * \Phi_{l-1}(x_i) \right\|_2^2
\end{equation}
where $l$ is the index of the convolutional layer to be approximated and $\Phi_{l}(x_i)$ is the evaluation of the CNN up to and including layer $l$ on data sample $x_i \in X$ where $X$ is the set of training examples. This optimization can be done quite elegantly by simply mirroring the CNN with the un-optimized separable basis layers, and training only the approximation layer by back-propagating the $L_2$ error between the output of the original layer and the output of the approximation layer (see Fig.~\ref{fig:reconnet}). This is done layer by layer.

There are two main advantages of this method for optimization of the approximation schemes. The first is that the approximation is conditioned on the manifold of the training data -- original filter dimensions that are not relevant or redundant in the context of the training data will by ignored by minimizing data reconstruction error, but will still be penalised by minimizing filter reconstruction error (Sec.~\ref{sec:filterrecon}) and therefore uselessly using up model capacity. Secondly, stacks of approximated layers can be learnt to incorporate the approximation error of previous layers by feeding the data through the approximated net up to layer $l$ rather than the original net up to layer $l$ (see Fig.~\ref{fig:reconnet} (b)). This additionally means that all the approximation layers could be optimized jointly with back-propagation.

An obvious alternative optimization strategy would be to replace the original convolutional layers with the un-optimized approximation layers and train just those layers by back-propagating the classification error of the approximated CNN. However, this does not actually result in better classification accuracy than doing $L_2$ data reconstruction optimization -- in practice, optimizing the separable basis within the full network leads to overfitting of the training data, and attempts to minimize this overfitting through regularization methods like dropout~\cite{Hinton12} lead to under-fitting, most likely due to the fact that we are already trying to heavily approximate our original filters. However, this is an area that needs to be investigated in more detail. 

\vspace{-2mm}
\section{Experiments \& Results}\label{sec:exp}
In this section we demonstrate the application of both proposed filter approximation schemes and show that we can achieve large speedups with a very small drop in accuracy. We use a pre-trained CNN that performs case-insensitive character classification of scene text. Character classification is an essential part of many text spotting pipelines such as~\cite{QuackText09,Posner10,Yang12,NeumannText10,Neumann-icdar2011,Neumann12,Wang11,Neumann13,Alsharif13,Bissacco2013}.

We first give the details of the base CNN model used for character classification which will be subject to speedup approximations. The optimization processes and how we attain the approximations of Scheme 1 \& 2 to this model are given, and finally we discuss the results of the separable basis approximation methods on accuracy and inference time of the model.

\paragraph{Test Model.}
For scene character classification, we use a four layer CNN with a softmax output. The CNN outputs a probability distribution $p(c|x)$ over an alphabet $C$ which includes all 26 letters and 10 digits, as well as a noise/background (no-text) class, with $x$ being a grey input image patch of size $24 \times 24$ pixels, which has been zero-centred and normalized by subtracting the patch mean and dividing by the standard deviation. The non-linearity used between convolutional layers is maxout~\cite{Goodfellow13maxout} which amounts to taking the maximum response over a number of linear models \eg the maxout of two feature channels $z^1_i$ and $z^2_i$ is simply their pointwise maximum: $h_i(z_i(u,v))= \max\{z^1_i(u,v),z^2_i(u,v)\}$. Table~\ref{table:model} gives the details of the layers for the model used, which is connected in the linear arrangement Conv1-Conv2-Conv3-Conv4-Softmax.

\begin{table}[t]
\begin{center}\footnotesize
\begin{tabular}{|c|c|c|c|c|c|c|} \hline
Layer name & Filter size & In channels & Out channels & Filters & Maxout groups & Time \\
\hline\hline
Conv1 & $9 \times 9$ & 1 & 48 & 96 & 2 & 0.473ms (8.3\%)\\
\hline
Conv2 & $9 \times 9$ & 48 & 64 & 128 & 2 & 3.008ms (52.9\%)\\
\hline
Conv3 & $8 \times 8$ & 64 & 128 & 512 & 4 & 2.160ms (38.0\%)\\
\hline
Conv4 & $1 \times 1$ & 128 & 37 & 148 & 4 & 0.041ms (0.7\%)\\
\hline
Softmax & - & 37 & 37 & - & - & 0.004ms (0.1\%)\\
\hline
\end{tabular}
\end{center}\vspace{-0.5em}
\caption{\small The details of the layers in the CNN used with the forward pass timings of each layer.}
\label{table:model}
\end{table}

\paragraph{Datasets \& Evaluation.}
The training dataset consists of 163,222 collected character samples
from a number of scene text and synthesized character
datasets~\cite{icdar2003dataset,ICDAR2005,ICDAR11,ICDAR2013,kaistdataset,deCampos09,Wang12}. The test set is the collection of 5379 cropped characters from the ICDAR 2003 training set after removing all non-alphanumeric characters as in~\cite{Wang11,Alsharif13}. We evaluate the case-insensitive accuracy of the classifier, ignoring the background class. The Test Model achieves state-of-the-art results of 91.3\%
accuracy compared to the next best result of 89.8\%~\cite{Alsharif13}.

\paragraph{Implementation Details.}
The CNN framework we use is the CPU implementation of
\texttt{Caffe}~\cite{Jia13caffe}, where convolutions are performed by
constructing a matrix of filter windows of the input, \texttt{im2col},
and using BLAS for the matrix-matrix multiplication between the
filters and data windows. We found this to be the fastest CPU CNN implementation attainable. CNN training is done with SGD with momentum of 0.9
and weight decay of 0.0005.  Dropout of 0.5 is used on all layers except Conv1
to regularize the weights, and the learning rate is adaptively reduced
during the course of training.

For filter reconstruction optimization, we optimize a separable basis
until a stable minimum of reconstruction error is reached. For data
reconstruction optimization, we optimize each approximated layer in
turn, and can incorporate a fine-tuning with joint
optimization.



\begin{figure}[t]
\centering
\begin{tabular}{cccc}
\includegraphics[width=0.22\textwidth]{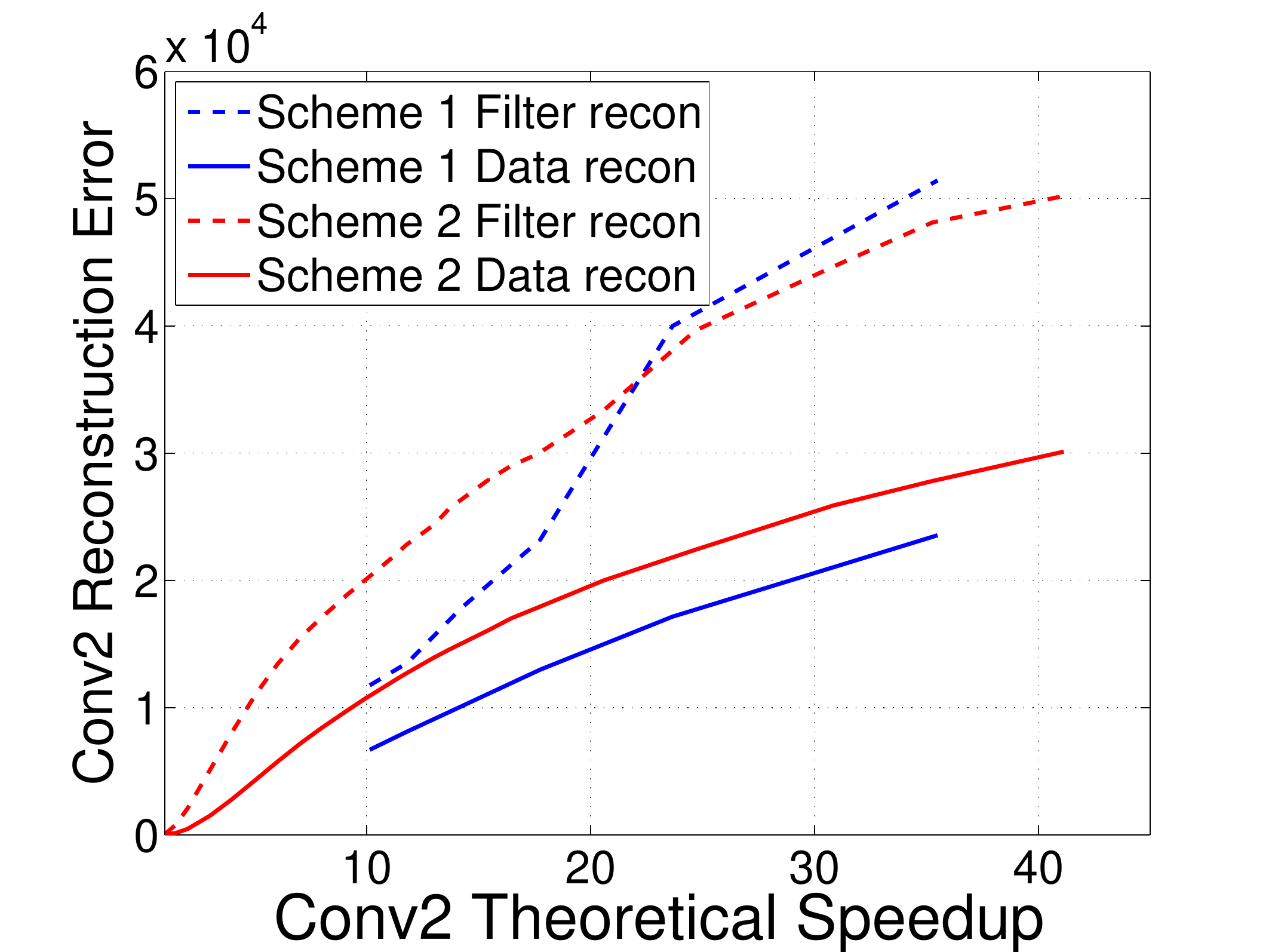}&
\includegraphics[width=0.22\textwidth]{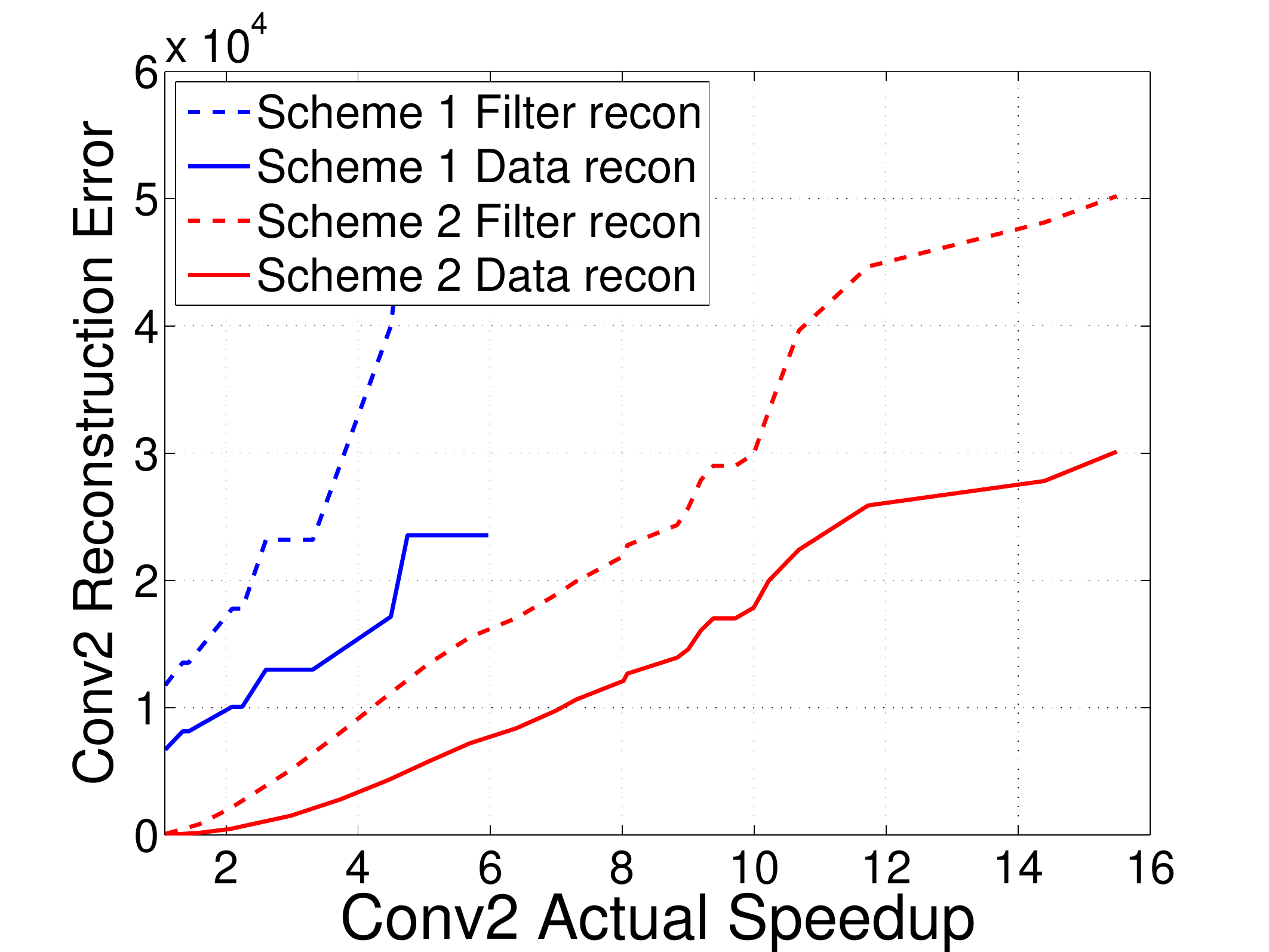}&
\includegraphics[width=0.22\textwidth]{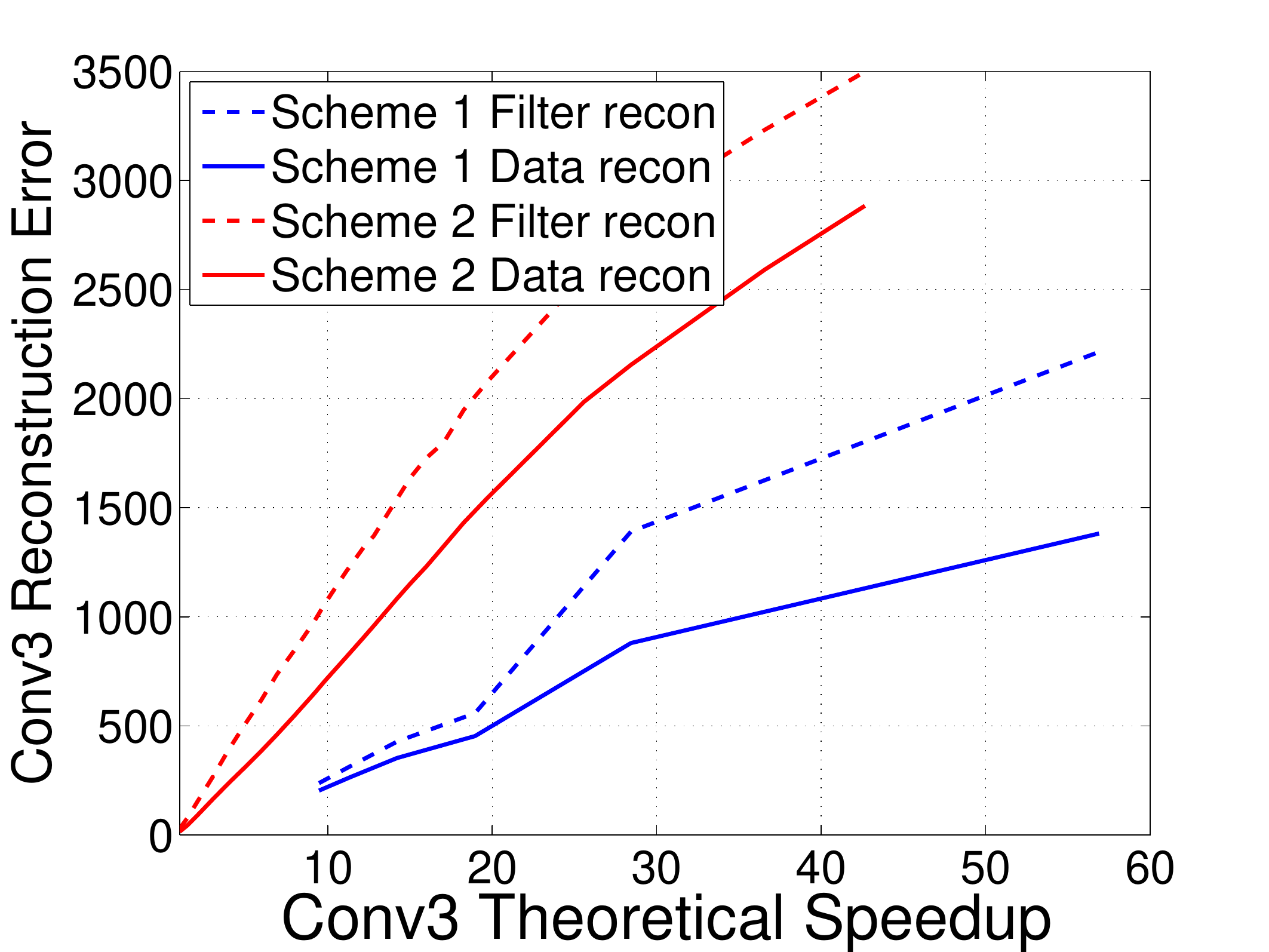}&
\includegraphics[width=0.22\textwidth]{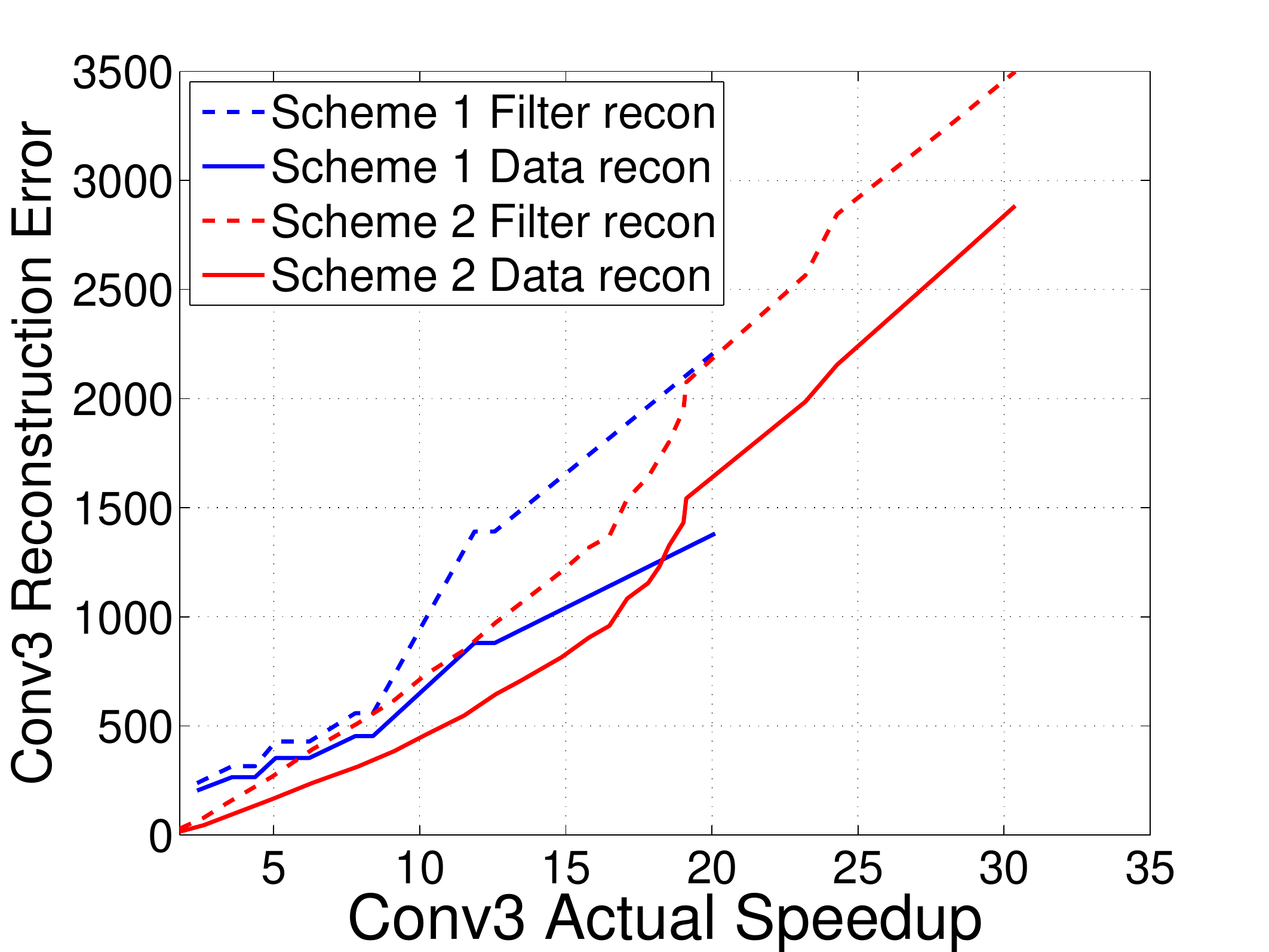}
\end{tabular}
\caption{\small Reconstruction error for the theoretical and actual attained speedups on test data for Conv2 \& Conv3. We do not go below 10$\times$ theoretical speedup for Scheme 1 as computation takes too long.}
\label{fig:layererror}
\end{figure}

\begin{figure}[t]
\vspace{-1em}
\hfill
\includegraphics[width=0.15\textwidth,trim=0 0 0 0]{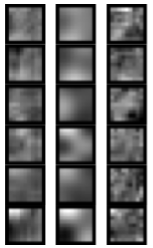}(a)\hfill
\includegraphics[width=0.32\textwidth,trim=0 0 40pt 0]{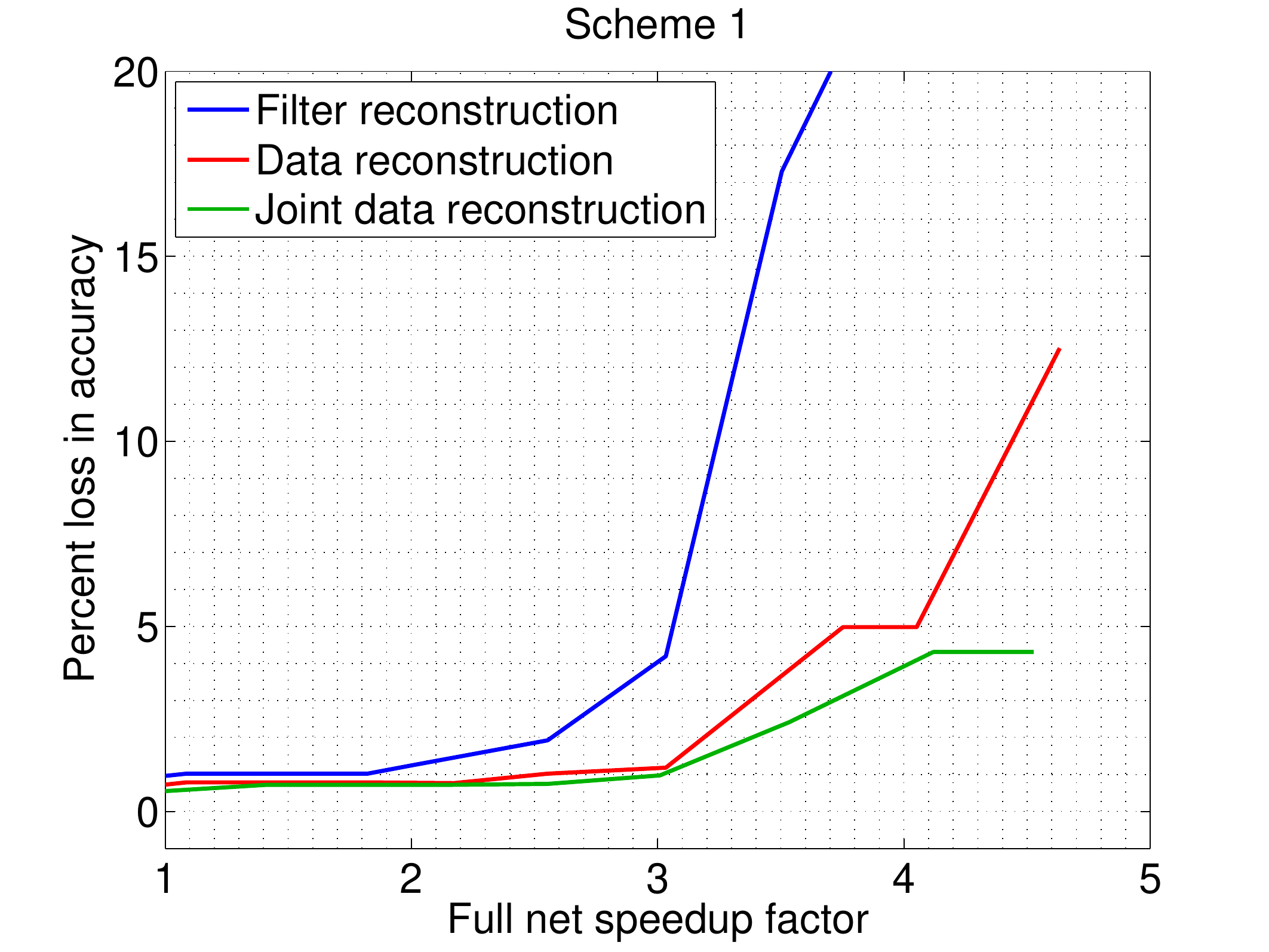}(b)\hfill
\includegraphics[width=0.32\textwidth,trim=0 0 40pt 0]{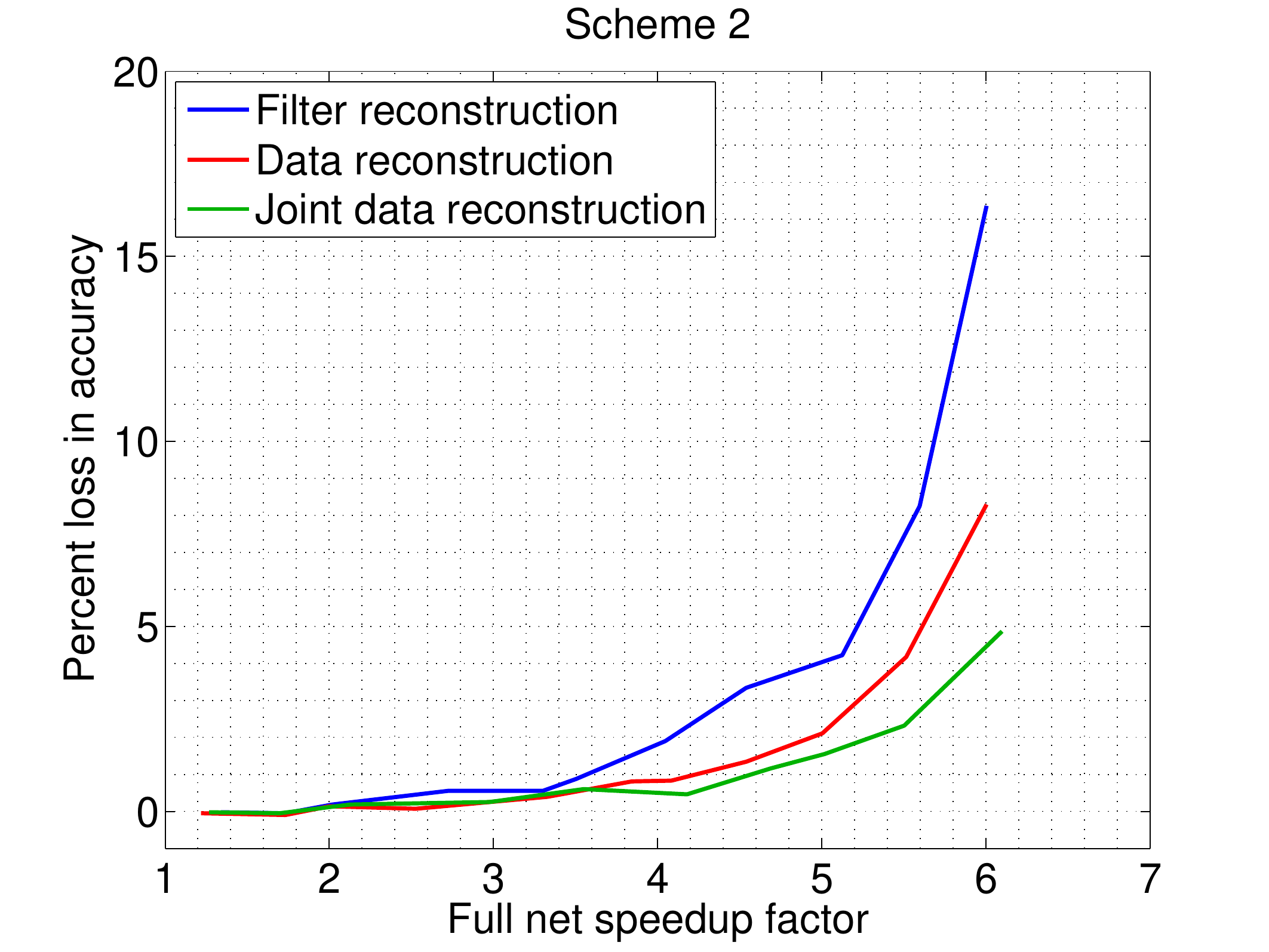}(c)\hfill\mbox{}
%
\caption{\small (a) 
A selection of Conv2 filters from the original CNN (left), and the
reconstructed versions under Scheme 1 (centre) and Scheme 2 (right),
where both schemes have the same model capacity corresponding to 10x
theoretical speedup. Visually the approximated filters look very different with Scheme 1 naturally smoothing the repesentation, but both still achieve good accuracy. (b-c) The percent loss in performance as a result
of the speedups attained with Scheme 1~(b) and Scheme 2~(c).}
\label{fig:neterror}
\end{figure}

\begin{figure}[t]\centering
\includegraphics[width=0.90\textwidth]{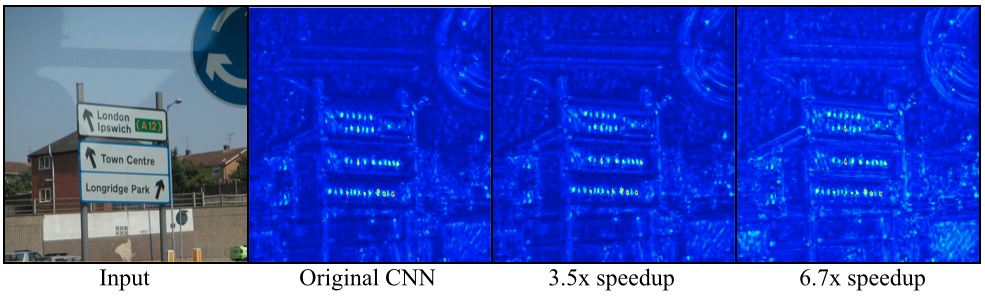}\vspace{-1em}
\caption{\small Text spotting using the CNN character classifier.
The maximum response map over the character classes of the CNN output
with Scheme 2 indicates the scene text positions. The approximations
have sufficient quality to locate the text, even at $6.7\times$
speedup.}
\label{fig:detmap}
\end{figure}


For the CNN presented, we only approximate layers Conv2 and Conv3. This is because layer Conv4 has a $1 \times 1$ filter size and so would not benefit much from our speedup schemes. We also don't approximate Conv1 due to the fact that it acts on raw pixels from natural images -- the filters in Conv1 are very different to those found in the rest of the network and experimentally we found that they cannot be approximated well by separable filters (also observed in~\cite{Denton2014}). Omitting layers Conv1 and Conv4 from the schemes does not change overall network speedup significantly, since Conv2 and Conv3 constitute 90\% of the overall network processing time, as shown in Table.~\ref{table:model}.

\paragraph{Layer-wise Performance.}
Fig.~\ref{fig:layererror} shows the output reconstruction error of each approximated layer with the test data. It is clear that the reconstruction error worsens as the speedup achieved increases, both theoretically and practically. As the reconstruction error is that of the test data features fed through the approximated layers, as expected the data reconstruction optimization scheme gives lower errors for the same speedup compared to the filter reconstruction. This generally holds even when completely random Gaussian noise data is fed through the approximated layers -- data from a completely different distribution to what the data optimization scheme has been trained on. 

Looking at the theoretical speedups possible in Fig.~\ref{fig:layererror}, Scheme 1 gives better reconstruction error to speedup ratio, suggesting that the Scheme 1 model is perhaps better suited for approximating convolutional layers. However, when the actual measured speedups are compared, Scheme 1 is actually slower than that of Scheme 2 for the same reconstruction error. This is due to the fact that the \texttt{Caffe} convolution routine is optimized for 3D convolution (summing over channels), so Scheme 2 requires only two \texttt{im2col} and BLAS calls. However, to implement Scheme 1 with \texttt{Caffe} style convolution involving per-channel convolution without channel summation, means that there are many more costly \texttt{im2col} and BLAS calls, thus slowing down the layer evaluation and negating the model approximation speedups. It is possible that using a different convolution routine with Scheme 1 will bring the actual timings closer to the theoretically achievable  timings.

\paragraph{Full Net Performance.}
Fig.~\ref{fig:neterror}~(b)~\&~(c) show the overall drop in accuracy
as the speedup of the end-to-end network increases under different
optimization strategies. Generally, joint data optimization of Conv2
and Conv3 improves final classification performance for a given
speedup. Under Scheme 2 we can achieve a 2.5$\times$ speedup with no loss in accuracy, and a 4.5$\times$ speedup with only a drop of 1\% in classification accuracy, giving
90.3\% accuracy -- still state-of-the-art for this benchmark. The 4.5$\times$
configuration is obtained by approximating the original 128 Conv2
filters with 31 horizontal filters followed by 128 vertical filters,
and the original 512 Conv3 filters with 26 horizontal filters followed
by 512 vertical filters.

%

This speedup is incredibly useful for sliding window schemes, allowing fast generation of, for example, detection maps such as the character detection map shown in Fig.~\ref{fig:detmap}. There is very little difference with even a 3.5$\times$ speedup, and when incorporated in to a full application pipeline, the speedup can be tuned to give an acceptable end pipeline result.

Comparing to an FFT based CNN~\cite{Mathieu13}, our method can actually give greater speedups. With the same layer setup (5$\times$5 kernel, $16\times 16 \times 256$ input, 384 filters), Scheme 2 gives an actual 2.4$\times$ speedup with 256 basis filters (which should result in no performance drop), compared to $2.2\times$ in~\cite{Mathieu13}. Comparing with~\cite{Denton2014}, simply doing a filter reconstruction approximation with Scheme 2 of the second layer of OverFeat~\cite{Sermanet13} gives a 2$\times$ theoretical speedup with only 0.5\% drop in top-5 classification accuracy on ImageNet, far better than the 1.2\% drop in accuracy for the same theoretical speedup reported in~\cite{Denton2014}. This accuracy should be further improved if data optimization is used.

\vspace{-2mm}
\section{Conclusions}\label{sec:conc}
In this paper we have shown that the redundancies in representation in CNN convolutional layers can be exploited by approximating a learnt full rank filter bank as combinations of a rank-1 filter basis. We presented two schemes to do this, with two optimization techniques for attaining the approximations. The resulting approximations require significantly less operations to compute, resulting in large speedups observed with a real CNN trained for scene text character recognition: a 4.5$\times$ speedup, only a drop of 1\% in classification accuracy.

In future work it would be interesting to experiment with other arrangements of separable filters in layers, \eg a horizontal basis layer, followed by a vertical basis layer, followed by a linear combination layer. Looking at the filter reconstructions of the two schemes in Fig.~\ref{fig:neterror}~(a), it is obvious that the two presented schemes act very differently, so the connection between different approximation structures could be explored. Also it should be further investigated whether these model approximations can be effectively taken advantage of during training, with low-rank filter layers being learnt in a discriminative manner.

\paragraph{Acknowledgements.} 
Funding for this research is provided by the EPSRC and ERC grant VisRec no. 228180.

\bibliography{shortstrings,vgg_local,vgg_other,max_bib}
\end{document}